%% file: main.tex
\documentclass[sigconf]{acmart}
\AtBeginDocument{%
  \providecommand\BibTeX{{%
    \normalfont B\kern-0.5em{\scshape i\kern-0.25em b}\kern-0.8em\TeX}}}

\setcopyright{acmcopyright}
\copyrightyear{2022}
\acmYear{2022}
\acmDOI{XXXXXXX.XXXXXXX}

\usepackage{multirow}
\usepackage{subcaption}
\usepackage{caption}
\usepackage{tikz}
\newtheorem{definition}{Definition}

\acmConference[GLB '22]{Workshop on Graph Learning Benchmarks, Web Conference}{April 25--29, 2022}{Virtual}
\acmPrice{15.00}
\acmISBN{978-1-4503-XXXX-X/18/06}

\input{sections/preamble_and_definitions}

\begin{document}

\title{EXPERT: Public Benchmarks for Dynamic Heterogeneous\\ Academic Graphs}% Datasets}%to Model Human Behavior}
%Domain-specific Academic Graph Datasets for Forecasting Global Expertise and Capability Development under Distribution Shift}

%
% The "author" command and its associated commands are used to define
% the authors and their affiliations.
% Of note is the shared affiliation of the first two authors, and the
% "authornote" and "authornotemark" commands
% used to denote shared contribution to the research.
% \author{Sameera Horawalavithana}
% \email{yasanka.horawalavithana@pnnl.gov}
% % \orcid{1234-5678-9012}
% \author{Ellyn Ayton}
% \email{ellyn.ayton@pnnl.gov}
% \author{Anastasiya Usenko}
% \email{anastasiya.usenko@pnnl.gov}
% \author{Robin Cosbey}
% \email{robin.cosbey@pnnl.gov}
% \author{Maria Glenski}
% \email{maria.glenski@pnnl.gov}
% \author{Svitlana Volkova}
% \email{svitlana.volkova@pnnl.gov}
% \affiliation{%
%   \institution{Pacific Northwest National Laboratory}
% %   \streetaddress{P.O. Box 1212}
%   \city{Richland}
%   \state{WA}
% %   \postcode{43017-6221}
% }
 
\notignore{
\author{Sameera Horawalavithana, 
        Ellyn Ayton,
        Anastasiya Usenko,
        Shivam Sharma, \\
        Jasmine Eshun, 
        Robin Cosbey,
        Maria Glenski, and
        Svitlana Volkova
        }
\affiliation{%
  \institution{Pacific Northwest National Laboratory
  \\
  ~\textit{firstname}.\textit{lastname}@pnnl.gov
  %yasanka.horawalavithana@pnnl.gov,
  }
%   \city{Richland, WA}
  \country{USA}
}
    %\email{yasanka.horawalavithana@pnnl.gov}
    %\email{firstname.lastname@pnnl.gov}
}
\ignore{ 
\author{Ellyn Ayton,
        Anastasiya Usenko,
        Robin Cosbey,
        Maria Glenski, and
        Svitlana Volkova
        }
\affiliation{%
  \institution{Pacific Northwest National Laboratory}
%   \city{Richland, WA}
}
\email{firstname.lastname@pnnl.gov}

\author{Svitlana Volkova
        }
\affiliation{%
  \institution{Visual Analytics \\ Pacific Northwest National Laboratory}
%   \city{Richland, WA}
}
\email{yasanka.horawalavithana@pnnl.gov}
}

\ignore{
\author{Sameera Horawalavithana}
\affiliation{%
  \institution{PNNL}
%   \city{Richland, WA}
}
\email{yasanka.horawalavithana@pnnl.gov}

\author{Ellyn Ayton}
\affiliation{%
  \institution{PNNL}
%   \city{Richland, WA}
}
\email{ellyn.ayton@pnnl.gov}

\author{Anastasiya Usenko}
\affiliation{%
  \institution{PNNL}
%   \city{Richland, WA}
}
\email{anastasiya.usenko@pnnl.gov}

\author{Robin Cosbey}
\affiliation{%
  \institution{PNNL}
%   \city{Richland, WA}
}
\email{robin.cosbey@pnnl.gov}

\author{Maria Glenski}
\affiliation{%
  \institution{PNNL}
%   \city{Richland, WA}
}
\email{maria.glenski@pnnl.gov}

\author{Svitlana Volkova}
\affiliation{%
  \institution{PNNL}
%   \city{Richland, WA}
}
\email{svitlana.volkova@pnnl.gov}
}

% %%
% %% By default, the full list of authors will be used in the page
% %% headers. Often, this list is too long, and will overlap
% %% other information printed in the page headers. This command allows
% %% the author to define a more concise list
% %% of authors' names for this purpose.
\renewcommand{\shortauthors}{Horawalavithana et al.}

\begin{abstract}
\input{sections/abstract}
\end{abstract}



\ignore{
}


\maketitle

\input{sections/introduction}

\input{sections/related}
\input{sections/processing}

\input{sections/tasks}

\input{sections/discussion}
\begin{acks}
This material is based on work funded by the United States Department of Energy (DOE) National Nuclear Security Administration (NNSA) Office of Defense Nuclear Nonproliferation Research and Development (DNN R\&D) Next-Generation AI research portfolio and Pacific Northwest National Laboratory, which is operated by Battelle Memorial Institute for the U.S. Department of Energy under contract DE-AC05-76RLO1830. Any opinions, findings, and conclusions or recommendations expressed in this material are those of the author(s) and do not necessarily reflect the views of the United States Government or any agency thereof. 
Authors thank Joonseok Kim and Sannisth Soni for their help with preparing the datasets.
\end{acks}

\bibliographystyle{ACM-Reference-Format}
\bibliography{bib-v1}

% %%
% %% If your work has an appendix, this is the place to put it.
% \appendix









\end{document}

%% file: sections/preamble_and_definitions.tex
\newcommand{\eg}{\textit{e.g.,~}}
\newcommand{\ie}{\textit{i.e.,~}} 
\newcommand{\ignore}[1]{ } 
\newcommand{\notignore}[1]{#1}

\usepackage{array}
\newcolumntype{H}{>{\setbox0=\hbox\bgroup}c<{\egroup}@{}}
\newcolumntype{h}{>{\setbox0=\hbox\bgroup}c<{\egroup}@{}}

\newcommand{\dhgraphs}{dynamic heterogeneous academic graphs~}

\usepackage{pgfplots}  
\usepackage{tikz}
\pgfplotsset{compat=1.16}

%% file: sections/abstract.tex
Machine learning models that learn from dynamic graphs face nontrivial challenges in learning and inference as both nodes and edges change over time. The existing large-scale graph benchmark datasets that are widely used by the community primarily focus on homogeneous node and edge attributes and are static. In this work, we present a variety of large scale, \dhgraphs to test the effectiveness of models developed for multi-step graph forecasting tasks. Our novel datasets cover both context and content information extracted from scientific publications across two  communities – Artificial Intelligence (AI) and Nuclear Nonproliferation (NN). In addition, 
% we evaluate predictive performance of graph forecasting models and 
we propose a systematic approach to improve the existing evaluation procedures used in the graph forecasting models.
%Data and code are available via 
%\url{https://github.com/pnnl/EXPERT}.

%% file: sections/introduction.tex
\section{Introduction}
% Graphs arise naturally in many real-world applications including social networks, recommender systems, ontologies, biology, and computational finance. 
% Traditionally, machine learning models have been designed for static graphs. 
% However, 
% Many applications involve evolving graphs. This introduces important challenges for learning and inference since nodes, attributes, and edges change over time.
% 
The adoption of several benchmark datasets by the graph-based learning community has spurred many research challenges and opportunities in large-scale graph modeling~\cite{hu2020open} and  out-of-distribution generalization~\cite{hu2021ogb,lim2021new}. 
These benchmarks are useful across inference tasks and domains to understand the limitations of existing Graph Neural Network (GNN) models or to develop new GNN models for state-of-the-art graph-based learning tasks, \eg link prediction. For example, the top-ranked solutions of the 2021 KDD cup developed large and deep GNN models to predict primary subject areas of Arxiv papers released in the Open Graph Benchmark (OGB)~\cite{hu2021ogb}.
% For example, the winning solution at KDD Cub 2021 GNN models are indeed large and deep; the number of learnable
% 347 parameters (single model) ranges from 50M up to 450M,
%  often leveraged
Another benchmark study revealed the limitations of existing GNN methods in non-homophilous graphs~\cite{lim2021new}.
% models perform poorly in non-homophilous graph settings thorough another benchmark study~\cite{lim2021new}.
% as a strong inductive bias is homophily, which means that connected
% nodes tend to be similar in certain attributes
%
Although most of the existing large-scale graph benchmark datasets focus on static homogeneous graphs~\cite{hu2021ogb,lim2021new}, many real-world problems involve dynamic graphs where heterogeneous nodes and edges change over time.
The evolving nature of dynamic graphs requires handling new, previously unseen nodes as well as capturing temporal patterns. Several promising graph-based learning frameworks have been developed for dynamic heterogeneous graphs~\cite{xu2020inductive,rossi2020temporal,jin2019recurrent}. 
The proposed approaches learn time-aware node embeddings to represent the evolving topological structures.
However, these methods are yet to be tested on standard large-scale dynamic graph benchmark datasets.
% 
% adopted in the graph machine learning community.
%
Scholarly publications present an avenue for studying dynamic heterogeneous graphs. Large-scale \dhgraphs capture scientific knowledge development and collaboration patterns across disciplines, \eg artificial intelligence, natural language processing, and can provide new insights on how careers evolve, how collaborations drive scientific discovery, and how scientific progress emerges~\cite{wang2021science}. 

% Is there a pattern underlying the way we collaborate?
% show that while some scientific collaborations are impactful and long-lasting, others fail abruptly. 
% For example, highly collaborative individuals are somewhat common in biology, but are especially rare in mathematics~\cite{wang2021science}.
% % diversity within a scientific team promotes the effectiveness of the team, diversity, including nationality, ethnicity, institution, gender, academic age and disciplinary backgrounds. 

There are two contributions of this work: (1) the public release seven novel dynamic, heterogeneous academic
graph benchmark datasets for two research communities (artificial intelligence and nuclear nonproliferation); and (2) standardized evaluation procedures for forecasting on dynamic, heterogeneous context graphs. We investigate and draw novel insights about performance evaluation for graph forecasting tasks using a systematic approach analyzing the complexity of both transductive and inductive predictions.

%% file: sections/related.tex
\section{Related Work}%Public Academic Graph Benchmarks}
In this section, we outline related work on academic graph benchmark datasets.
Existing graph benchmarks primarily focus on either node classification or link prediction tasks.
Cora~\cite{getoor2005link} and CiteSeer~\cite{getoor2005link} datasets are widely used~\cite{yang2016revisiting} for node classification, where the task is to predict the missing subject area of a paper represented as a node in the graph.
However, these datasets are very small ($<4K$ nodes) which makes them unsuitable for evaluation of graph-based ML models; moreover, there have been several issues reported about data quality~\cite{dwivedi2020benchmarking,hu2020open}.
Open Graph Benchmark (OGB)~\cite{hu2020open} released three academic graph benchmark datasets (\eg ogbn-arxiv, ogbn-papers100M, ogbn-mag) for node classification tasks.
OGB datasets are extracted from the Microsoft Academic Graph (MAG)~\cite{wang2020microsoft}.
Several recent works show the usefulness of the OGB datasets for large-scale graph learning~\cite{hu2021ogb}.
In link prediction tasks, evaluation is focused on predicting missing edges. DBLP~\cite{yang2015defining} and ogbl-citation2~\cite{hu2020open} are two commonly used benchmarks for link prediction in static citation networks.  Similarly, HEP-PH~\cite{gehrke2003overview} is a benchmark for link prediction in a citation network, but considers a dynamic graph setting.
However, these datasets are not suited for evaluation on edge forecasting tasks on \textit{heterogeneous} graphs.
As shown in Table~\ref{tab:benchmakr_table}, most existing large-scale graph benchmarks focus on static, homogeneous graphs and are mainly co-citation and collaboration networks.
The most similar to our work is the \textit{ogbn-mag} benchmark released as a part of the Open Graph Benchmark (OGB)~\cite{hu2020open}.
This heterogeneous graph dataset includes several types of nodes (authors, institutions, papers and topics) with multiple edge types -- affiliations, authorships, citations and paper-topic relationships.
The \textit{ogbn-mag} prediction task is to predict the missing venue (conference or journal).
While this dataset is useful for the development of heterogeneous graph-based models, it does not take into account the timestamped edges.
Unlike any previous work, we present a novel \dhgraphs dataset representing seven data sources across two domains.% representing two different scientific communities (AI and nuclear nonproliferation).

% Cora~\cite{getoor2005link} & 2,708 & 5,429
% CiteSeer~\cite{getoor2005link} & 3,312 & 4,732
% ogbn-arxiv~\cite{hu2020open} & 169,343 & 1,166,243
% ogbn-papers100M~\cite{hu2020open} & 111,059,956 & 1,615,685,872
% ogbl-collab~\cite{hu2020open} & 235,868 & 1,285,465
% ogbl-citation2~\cite{hu2020open} & 2,927,963 & 30,561,187
% DBLP~\cite{yang2015defining} & 317,080 & 1,049,866
% ogbn-mag~\cite{hu2020open} & 1,939,743 & 21,111,007
% HEP-PH~\cite{gehrke2003overview} & 34,546 & 421,578

\begin{table}[htbp]
    \small
    \centering
      \caption{Existing benchmarks for academic graph datasets, including whether graphs are dynamic (D). We use OGB statistics reported in the original paper~\cite{hu2020open}.}
    \label{tab:benchmakr_table}
    \setlength{\tabcolsep}{2pt} %default is 6
    \begin{tabular}{l|p{10mm}|p{10mm}|c|p{15mm}|p{15mm}}
          Benchmark & \# Nodes & \# Edges & D & Edge Types & Pred. Task   \\ \hline \hline
         
        Cora~\cite{getoor2005link} & 2.7K & 5.4K & -- & citation &  subject areas \\ \hline

         CiteSeer~\cite{getoor2005link} & 3.3K & 4.7K & -- & citation &  subject areas \\ \hline
         
         ogbn-arxiv~\cite{hu2020open} & 0.2M & 1.1M & -- & citation &  subject areas \\ \hline
         
         ogbn-papers100M~\cite{hu2020open} & 111M & 1.6B & --  & citation &  subject areas  \\ \hline
         
         ogbl-collab~\cite{hu2020open} & 0.2M & 1.3M & --  & collaboration & collaborations  \\ \hline
         
         ogbl-citation2~\cite{hu2020open} & 2.9M & 30.6M & --  & citation &  citations  \\ \hline
         
         DBLP~\cite{yang2015defining} & 0.3M & 1.1M & -- & citation & citations \\ \hline
         
        %  mag240M~\cite{hu2021ogb} & 1,939,743 & 21,111,007 & static  & \begin{tabular}{@{}c@{}}affiliations, authorship,\\ citations, topics \end{tabular} & venue \\ \hline
        
         ogbn-mag~\cite{hu2020open} & 1.9M & 21M & --  & affiliations, authorship, citations, topics  & venue \\ \hline
         
        %  HEP-TH~\cite{gehrke2003overview} & 7,980 & 21,036 & dynamic  & citation & citation \\ \hline
        
        HEP-PH~\cite{gehrke2003overview} & 34K & 0.4M & \checkmark  & citation & citations \\ \hline
         
         \textbf{Our Work*} & 3.5M & 34M & \checkmark  & collaboration, partnership,  expertise  & collaboration, partnership, expertise  \\ \hline
    \end{tabular}
\end{table}

%% file: sections/processing.tex
\section{Graph Dataset Construction}
We make available\footnote{All data is available from our GitHub and hosted by the Berkeley Data Cloud (\url{https://bdc.lbl.gov}) under the \textit{Global Expertise Forecasting} project} \dhgraphs for seven sources across two research communities --- artificial intelligence (AI) and nuclear nonproliferation.
Graphs are split temporally into train, validation, and test sets.
We reserve the last year(s) of data as the test set, depending on venue scale. The year prior to test is used for validation and remaining years for training. A summary of node and edge types are presented for each data split in Table~\ref{tab:data_char}.

%\subsection{Data Collection and Processing}
\paragraph{\textbf{Data Collection}} 
%We collected scientific publications across two research communities --- artificial intelligence (AI) and nuclear nonproliferation.  
%
AI papers were collected from the proceedings of four top-tier AI conferences: Association for Computational Linguistics (ACL), the International Conference on Machine Learning (ICML), the International Conference on Learning Representations (ICLR) and the Conference on Neural Information Processing Systems (NeurIPS). 
We downloaded publication PDFs, which we parsed using GROBID \cite{GROBID} and CERMINE \cite{tkaczyk2015cermine} to extract content as well as metadata (author affiliations, country affiliations, etc.). %Both methods encountered inaccuracies caused by irregular organizations within the PDFs, missing information, and parser errors.
To obtain focused benchmarks, we filtered publications using a set of keywords curated by subject matter experts (SMEs)\footnote{AI keywords: fair, 
ethic, 
translation model, 
machine translation, 
dialog, 
genetic algorithm, 
explanation, 
transfer learning, 
clustering, 
adversarial, 
nlg, 
sentiment, 
causal, 
reinforcement learning, 
transparent, 
summarization, 
question-answer, 
interpretability,  
language model, 
interpretable%\url{https://github.com/pnnl/EXPERT/graph_benchmark_AI.txt}
}.
Nuclear publication records and abstracts were collected from three sources: the Office of Scientific and Technical Information (OSTI), Web of Science (WoS), and the SCOPUS database using a set of nuclear keywords\footnote{Nuclear keywords from: \url{github.com/pnnl/expert/blob/master/expert/queries.py}} compiled with domain resources% %(IAEA Safety Glossary
\footnote{IAEA Safety Glossary:
\url{www-pub.iaea.org/MTCD/Publications/pdf/PUB1830_web.pdf}}
 %)
and manual specification from SMEs. 
%After identifying this set of English terms and queries, we translated the terms and queries into three other languages (Russian, Korean, and Arabic) to increase the multilingual coverage, focused on these languages.
False positives were removed using topic modeling~\cite{angelov2020top2vec} and SME verification of non-nuclear topics.% to remove. 

\begin{figure}[!t]
    \centering
  \includegraphics[width=.9\columnwidth]{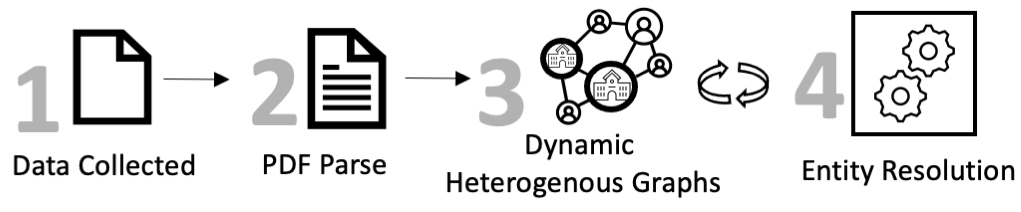} % scale=0.6
  \caption{Academic graph data pre-processing pipeline.}
%   \vspace{-0.5\baselinestretch}
  \label{fig:workflow}
\end{figure}

\begin{figure}[!t]
    \centering
    \includegraphics[width=.7\columnwidth]{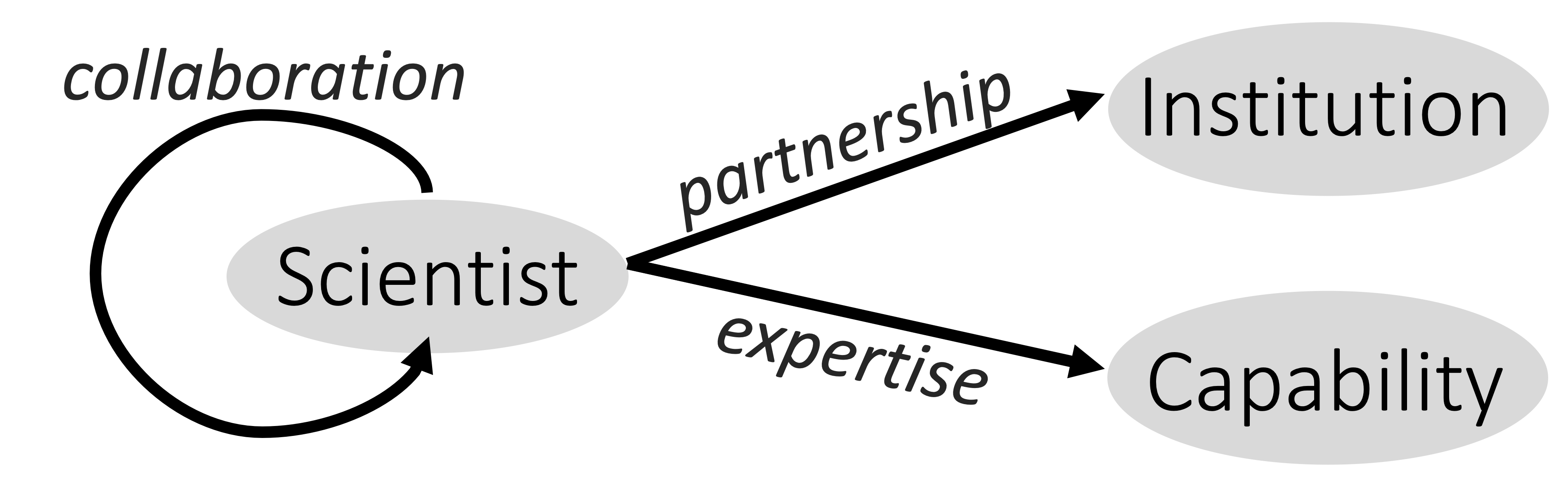} 
    %\vspace{-0.5\baselinestretch}
    \caption{Schema of our dynamic academic graphs.}
    %Relationships and entities in our dynamic graphs.}
    \label{fig:schema-graph}
\end{figure}

\paragraph{\textbf{Processing and Graph Construction}}
%\subsection{Data Processing Pipeline}
%Figure~\ref{fig:workflow} illustrates the data processing steps used to prepare each dataset. 
%First, we downloaded publication PDFs, in some instances (\ie OSTI) JSON records were available.
%We parsed the PDFs using GROBID \cite{GROBID} and CERMINE \cite{tkaczyk2015cermine} to extract content as well as metadata (author affiliations, country affiliations, etc.). 
%Both methods encountered inaccuracies caused by irregular organizations within the PDFs, missing information, and parser errors.
%
Using an in-house developed processing pipeline outlined in Figure~\ref{fig:workflow}, we processed publication meta-data (aka context) and 
%and publication full text (aka content) 
into \dhgraphs (schema in Figure~\ref{fig:schema-graph}). For every dataset, we performed entity resolution over all scientists and institutions.
%We define capabilities as the set of domain-specific keywords used to create the datasets. 
%
% We then performed entity resolution over all scientists and institutions. 
%
Given the many ways author and institution names can be represented (\eg Jane Doe and J. Doe), we created unique nodes for authors and affiliations in each paper and manually combined nodes if two entities were determined to be the same. 
When resolving nodes, we consider text similarity, the edit distance between the two names, and graph similarity of nodes' ego networks. 
For example, the two scientists, Jane Doe and J. Doe, have a high text similarity since their names are almost identical. 
These scientists are merged if they also have a high graph similarity score, \ie similar coauthors, capabilities and institutions. 
Leveraging both metrics prevents us from incorrectly merging scientists, \eg Jane Doe and John Doe.
We iteratively apply this process to the graphs until only node pairs with low similarity scores are returned. 
Nuclear datasets with >500K nodes were partially manually resolved due to size. 
In this case, we applied an automatic resolver on nodes with similarity scores above a threshold heuristically determined for each dataset.
%Scientists were mapped to their affiliated countries, using resolved links from scientist to institution to country.
%Finally, scientists and institutions are mapped to their affiliated countries.
Processing code is available on our GitHub\footnote{Data  processing code: \url{https://github.com/pnnl/EXPERT/tree/master/examples}}.

%\shnote{describe BDC and add the BDC link.}
\ignore{
\begin{table*}[htbp]
    \centering
    \caption{Summary of the %artificial intelligence and nuclear non-proliferation 
    \dhgraphs when split into train, test, and validation sets. % used in the experiments during training and evaluation. 
    % \shnote{Ana, can you update this table. Use the original non-filtered datasets for WoS, and SCOPUS. And ACL entire timeline.}
    }
    \label{tab:data_char}
    \begin{tabular}{|c|c|l|r|r|r|r|r|r|r|}
    \hline
         \multirow{2}{*}{\textit{}} & \multirow{2}{*}{\textit{Venue}} & \multirow{2}{*}{\textit{Data Split}} & \multirow{2}{*}{\textit{Time Range}} & \multicolumn{3}{c|}{\textit{\# Nodes}} & \multicolumn{3}{c|}{\textit{\# Edges}} \\ \cline{5-10}
         & & & & \#Scientists & \#Institutions & \#Capabilities & \#Collaborations & \#Partnerships & \#Expertise \\ \hline \hline
         %%AI
         \multirow{9}{*}{\rotatebox{90}{\textbf{AI Domain}}}
         & \multirow{3}{*}{\textbf{ACL}} & Training & 1965-2018 & 38.4K & 8K & 20 & 237.2K & 207.1K & 10.4K \\ %\cline{3-10}
          & & Validation & 2019 & 9.7K & 2K & 19 & 45.2K & 32K & 1.9K \\ %\cline{3-10}
          & & Testing & 2020-2021 & 8.4K & 1.9K & 20 & 34.2K & 28.9K & 1.1K \\ \cline{2-10}
         & \multirow{3}{*}{\textbf{ICML}} & Training & 2009-2019 & 9.2K & 493 & 20 & 5.9K & 20.7K & 12.6K \\ %\cline{3-10}
          & & Validation & 2020 & 4K & 281 & 19 & 2.3K & 8K & 4.6K \\ %\cline{3-10}
          & & Testing & 2021 & 4.6K & 301 & 18 & 2.5K & 10.1K & 5.2K \\ \cline{2-10}
         & \multirow{3}{*}{\textbf{ICLR}} & Training & 2016-2019 & 5.3K & 490 & 19 & 3.9K & 11.9K & 5.8K \\ %\cline{3-10}
          & & Validation & 2020 & 2,639 & 276 & 20 & 1.9K & 6.6K & 2.9K \\ %\cline{3-10}
          & & Testing & 2021 & 3.4K & 284 & 20 & 2.5K & 9K & 3.8K \\ \cline{2-10}
         & \multirow{3}{*}{\textbf{NeurIPS}} & Training & 1987-2018 & 22.2K & 1.4K & 20 & 12.7K & 31.1K & 22,619 \\ %\cline{3-10}
          & & Validation & 2019 & 4.6K & 454 & 19 & 2.9K & 9.3K & 4.8K \\ %\cline{3-10}
          & & Testing & 2020 & 8K & 572 & 20 & 4.9K & 16,962 & 8,570 \\ \hline
          %%Nuclear
         \multirow{9}{*}{\rotatebox{90}{\textbf{Nuclear Domain}} }
         & \multirow{3}{*}{\textbf{WoS}} & Training & 2015-2018 & 1.3M & 108.6K & 61 & 4.1M  & 10.9M & 156.5K \\ %\cline{3-10}
          & & Validation & 2019 & 449.1K & 48.6K & 47 & 1.28M & 3.1M & 37.9K \\ %\cline{3-10}
          & & Testing & 2020 & 311.2K & 34K & 47 & 867.6K & 2.08M & 23.2K \\ \cline{2-10}
         & \multirow{3}{*}{\textbf{SCOPUS}} & Training & 2015-2018 & 222.1K & 88.8K & 43 & 1.18M & 2.4M & 460.6K  \\ %\cline{3-10}
          & & Validation & 2019 & 91,614 & 34,385 & 42 & 387,267 & 748,045 & 143,513  \\ %\cline{3-10}
          & & Testing & 2020-2021 & 62,665 & 26,061 & 40 & 258,153 & 499,326 & 91,117  \\ \cline{2-10} 
         & \multirow{3}{*}{\textbf{OSTI}} & Training & 2015-2018 & 247,724 & 17,303 & 43 & 176,395 & 1,203,955 & 520,978 \\ %\cline{3-10}
          & & Validation & 2019 & 25,008 & 2,584 & 39 & 38,014 & 337,073 & 46,022 \\ %\cline{3-10}
          & & Testing & 2020 & 47,222 & 4,994 & 41 & 119,676 & 1,363,581 & 119,676 \\ \hline
    \end{tabular}
\end{table*}
}

\notignore{
\begin{table*}[htbp]
    \small
    \centering
    \caption{Characteristics of the %artificial intelligence and nuclear non-proliferation 
    datasets used in the experiments during training and evaluation. 
    % \shnote{Ana, can you update this table. Use the original non-filtered datasets for WoS, and SCOPUS. And ACL entire timeline.}
    }
    \label{tab:data_char} 
    \begin{tabular}{|c|c|l|r|r|r|r|r|r|r|}
    \hline
    \multirow{2}{*}{\textit{}} & \multirow{2}{*}{\textit{Graph}} & \multirow{2}{*}{\textit{Data Split}} & \multirow{2}{*}{\textit{Time Range}} & \multicolumn{3}{c|}{\textit{\# Nodes}} & \multicolumn{3}{c|}{\textit{\# Edges}} \\ \cline{5-10}
         & & & & \#Scientists & \#Institutions & \#Capabilities & \#Collaborations & \#Partnerships & \#Expertise \\ \hline \hline
         %%AI
         \multirow{9}{*}{\rotatebox{90}{\textbf{AI Domain}}}
         & \multirow{3}{*}{\textbf{ACL}} & Training & 1965-2018 & 38,436 & 7,979 & 20 & 237,174 & 207,096 & 10,430 \\ %\cline{3-10}
          & & Validation & 2019 & 9,691 & 1,966 & 19 & 45,231 & 31,989 & 1,941 \\ %\cline{3-10}
          & & Testing & 2020-2021 & 8,386 & 1,848 & 20 & 34,237 & 28,909 & 1,103 \\ \cline{2-10}
         & \multirow{3}{*}{\textbf{ICML}} & Training & 2009-2019 & 9,229 & 493 & 20 & 5,905 & 20,672 & 12,599 \\ %\cline{3-10}
          & & Validation & 2020 & 3,980 & 281 & 19 & 2,297 & 8,032 & 4,581 \\ %\cline{3-10}
          & & Testing & 2021 & 4,564 & 301 & 18 & 2,552 & 10,088 & 5,199 \\ \cline{2-10}
         & \multirow{3}{*}{\textbf{ICLR}} & Training & 2016-2019 & 5,272 & 490 & 19 & 3,900 & 11,934 & 5,797 \\ %\cline{3-10}
          & & Validation & 2020 & 2,639 & 276 & 20 & 1,925 & 6,559 & 2,916 \\ %\cline{3-10}
          & & Testing & 2021 & 3,407 & 284 & 20 & 2,498 & 9,006 & 3,881 \\ \cline{2-10}
         & \multirow{3}{*}{\textbf{NeurIPS}} & Training & 1987-2018 & 22,150 & 1,441 & 20 & 12,727 & 31,059 & 22,619 \\ %\cline{3-10}
          & & Validation & 2019 & 4,572 & 454 & 19 & 2,934 & 9,256 & 4,810 \\% \cline{3-10}
          & & Testing & 2020 & 7,968 & 572 & 20 & 4,918 & 16,962 & 8,570 \\ \hline
          %%Nuclear
         \multirow{9}{*}{\rotatebox{90}{\textbf{Nuclear Domain}}}
         & \multirow{3}{*}{\textbf{WoS}} & Training & 2015-2018 & 1,309,530 & 108,562 & 61 & 4,075,741  & 10,903,275 & 156,476 \\ %\cline{3-10}
          & & Validation & 2019 & 449,094 & 48,556 & 47 & 1,280,842 & 3,026,751 & 37,868 \\ %\cline{3-10}
          & & Testing & 2020 & 311,206 & 34,021 & 47 & 867,578 & 2,076,699 & 23,186 \\ \cline{2-10}
         & \multirow{3}{*}{\textbf{SCOPUS}} & Training & 2015-2018 & 222,051 & 88,758 & 43 & 1,182,054 & 2,399,450 & 460,579  \\ %\cline{3-10}
          & & Validation & 2019 & 91,614 & 34,385 & 42 & 387,267 & 748,045 & 143,513  \\% \cline{3-10}
          & & Testing & 2020-2021 & 62,665 & 26,061 & 40 & 258,153 & 499,326 & 91,117  \\ \cline{2-10} 
         & \multirow{3}{*}{\textbf{OSTI}} & Training & 2015-2018 & 247,724 & 17,303 & 43 & 176,395 & 1,203,955 & 520,978 \\ %\cline{3-10}
          & & Validation & 2019 & 25,008 & 2,584 & 39 & 38,014 & 337,073 & 46,022 \\% \cline{3-10}
          & & Testing & 2020 & 47,222 & 4,994 & 41 & 119,676 & 1,363,581 & 119,676 \\ \hline
    \end{tabular}
\end{table*}
}

% \subsection{Distribution Shifts due to Emerging Topics, New Collaboration and Partnerships}
% \shnote{Temporal shift, sub-population shifts.}

%% file: sections/tasks.tex
\section{Task Formulation}
We introduce several questions of interest that can be answered by studying our proposed dynamic heterogeneous academic graphs.
%We group them into three categories: collaboration, partnership, and capability evolution. %authorship. %patterns.

\paragraph{\textbf{Collaborations}}
Scientific publications are authored by individuals or teams of scientists.
Previous work~\cite{wang2021science} has shown that team-authored publications are more popular in terms of citations than single-authored publications. Questions focus on underlying patterns of collaboration:
%
%Is there a pattern underlying the way we collaborate? 
%Which kinds of scientists are most/least willing to collaborate?
% Who is the scientist any given scientist will collaborate with next?
Are there persistent groups of scientists who collaborate repeatedly? 
Do veteran scientists collaborate with early career or veteran scientists? 
Do collaborations occur within tightly connected groups of scientists?

\paragraph{\textbf{Partnerships}}
%Team work can improve productivity and result in groundbreaking scientific research and innovation.
Scientists may collaborate with other scientists from the same institution or across multiple.
What drives such multi-institutional partnerships?
Are researchers at elite universities more likely to collaborate with scientists at other elite universities?
To what extent do scientists collaborate internationally? % or domestically?
Are papers authored by international groups of scientists more likely to be published in high-impact journals?
% When a scientist is going to partner with another institution for the first time?
% Who are the scientists going to publish in a specific venue?
% Which institutions will publish from a given country?

\paragraph{\textbf{Capability Evolution}}
%Collaboration and multi-institutional partnerships are important factors for advancing scientific knowledge generation and dissemination. 
Teams of scientists produce diverse but specialized capabilities in comparison to what any individual collaborator could produce ~\cite{wang2021science}.
% large teams are better at further developing existing science and technology, while small teams disrupt science by suggesting new problems and opening up novel opportunities.
Are there differences in the topics that scientists tend to tackle?
Are scientists going to adopt the most recent and emerging capabilities?
Which scientists will disrupt science by suggesting new tasks and opening up novel opportunities?
How often do scientists generate more theoretical innovations in contrast to empirical analyses?
% What is the next capability a scientist will publish on?
% How many number of scientists will publish on a given capability?
% How often scientists jump between  same capability over time?

% We break the temporal link prediction problem in the the \dhgraphs into multiple sub-problems in the transductive and inductive settings.
% We introduce numerous problems of interests that can be defined by the temporal link prediction tasks.
% In a nutshell, we can predict relationships between a pair of nodes in multiple types (scientist, institution, and capability) to answer following research questions. 
% \begin{enumerate}
%     \item Can we predict whether two scientists collaborate repeatedly?
%     \item Would an incumbent scientist collaborate with a new scientist or another incumbent?
%     \item Who are the scientists going to adopt most recent and emerging capabilities?
%     \item Are the scientists going to build expertise on the same capability over time?
%     \item When a scientist is going to partner with another institution for the first time?
% \end{enumerate}
% We understand few limitations in the way the temporal link prediction is formulated and presented in previous works.

\subsection{Multi-step Link Prediction Task}
We consider \dhgraphs (G) consisting of scientists, institutions, and capabilities as nodes (N).
A pair of nodes is connected at a timestamp (t), by a directed edge which is denoted by a quadruple (\textit{$N_s$}, r, \textit{$N_o$}, t). 
Edges are of multiple types (r) such as collaboration (\textit{scientist-to-scientist}), partnership (\textit{scientist-to-institution}) and proficiency (\textit{scientist-to-capability}).
An ordered sequence of quadruples represents the dynamic heterogeneous graph.
In contrast to predicting missing edges in a static graph (\textit{interpolation}), we need to predict the future edges in a dynamic graph (\textit{extrapolation}).
As these edges occur over multiple time stamps in the future, we treat the prediction task as a multi-step inference task
% (see Figure~\ref{fig:forecasting_task}).
(see Definition~\ref{def:problem}).
Thus, we need to develop methods that can extrapolate the graph structure over future timestamps~\cite{jin2019recurrent}.
In our use case, such predictions are useful to forecast emerging science trends in terms of global expertise and capability development.

\ignore{
\begin{figure}[htbp]
    \centering
    \includegraphics[scale=0.25]{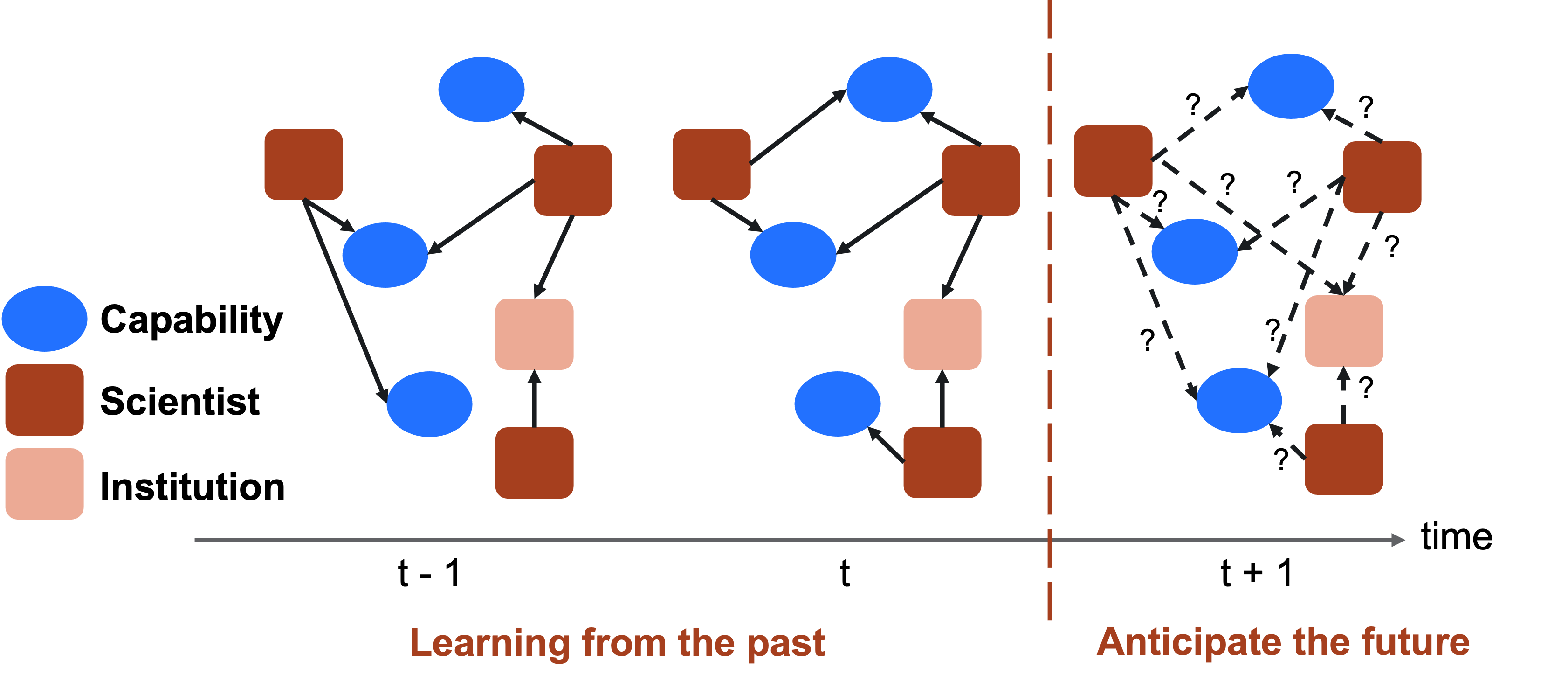}
    \caption{Multi-step heterogeneous link prediction.}
    \label{fig:forecasting_task}
\end{figure}
}

% \paragraph{\textbf{Problem Definition}}
\begin{definition}
\label{def:problem}
Given a graph ($G_{T}$) that represents the ordered sequence of quadruples until time $T$, the task is to forecast the graph ($G_{T:T+m}$) over multiple future time steps (m).
$G_{T}$ can be represented as \textit{discrete-time dynamic graphs} (\eg sequences of static graph snapshots) and~\textit{continuous-time dynamic graphs} (\eg timed lists of edges).
\end{definition}

\subsection{Task Complexity}
\label{sec:problem_complexity}
\begin{figure}[htbp]
\small
\centering
\begin{tikzpicture}[sibling distance=10em,level distance = 2.5em,
  every node/.style = {shape=rectangle, rounded corners,
    draw, align=center,
    top color=white, bottom color=blue!20}]]
  \node[bottom color=black!20] {Temporal Link Prediction}
    child { node[bottom color=orange!20, xshift=-3em] {Transductive} 
         child { node[yshift=0em] {Seen-to-Seen}
            child { node {First-time} }
            child { node {Repeated} }
         }
    }
    child { node[bottom color=orange!20,xshift=1em] {Inductive} 
        child { node[xshift=5em] {Unseen-to-Seen} }
        child { node[xshift=4em] {Seen-to-Unseen} }
         child { node[yshift=-2em,xshift=-10em] {Unseen-to-Unseen} }
    };
\end{tikzpicture}
%\vspace{-0.5\baselinestretch}
\caption{Forecasting Task Taxonomy. 
% Nodes in the test set are labeled seen or unseen based on whether they appeared in the train set, \eg  X is labeled as "unseen" in test if X was not in train.
}
\label{fig:task_taxonomy}
\end{figure}
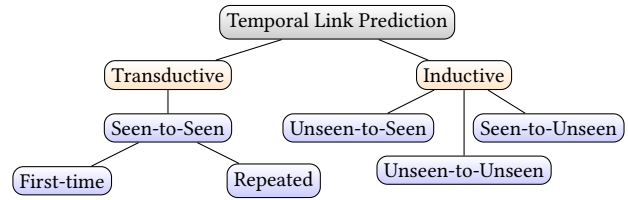

In this section, we discuss temporal link prediction as transductive and inductive tasks as illustrated in the taxonomy in Figure~\ref{fig:task_taxonomy}. 
% \eanote{Can we give this figure a caption and label?}
Our objective is to understand the complexity of different setups. %the graph forecasting via the temporal link prediction task.
% independent of the modeling ranking and classification approaches.

% \vspace{1em}
% \vspace{1em}

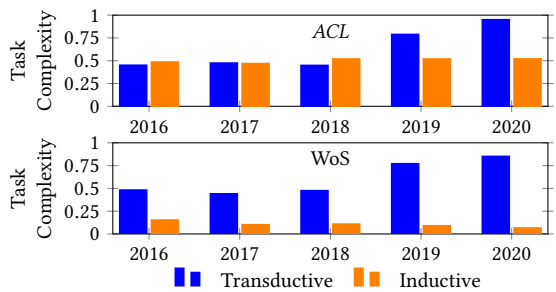
\begin{figure}[htbp]
% \begin{subfigure}{.48\textwidth}
%   \centering
  \input{images/complexity_plot}
%   \includegraphics[width=.98\linewidth]{images/acl_task_complexity.png}
%   \caption{ACL}
%   \label{fig:acl_task_complexity}
% \end{subfigure}
% \begin{subfigure}{.48\textwidth}
%   \centering
%   \includegraphics[width=.98\linewidth]{images/wos_task_complexity.png}
%   \caption{WoS}
%   \label{fig:wos_task_complexity}
% \end{subfigure}
\caption{Transductive and inductive task complexity. 
% We split the dataset into multiple train and test non-overlapping subsets for this analysis, where the test subset covers the year cited in the x-axis and the corresponding training subset covers the records timestamped before the test year. The complexity of the transductive and inductive tasks is due to the distribution shift as observed in the training and test datasets.
}
% \vspace{-0.5pt}
\label{fig:task_complexity}
\end{figure}

%We label the nodes in the testing data as "seen" or "unseen" as they co-appear in the training and testing data. For example, we label a scientist in the test data as "unseen" when she was not seen in the training data.
We group the edges in the test data into multiple categories within transductive (test sets only include edges between nodes \textit{seen} in training) and inductive (there is at least one "unseen" node in the test edges that was not present in training data) settings.%based on the labels described in the taxonomy.
% as they fit into both transductive and inductive settings.
In the transductive setting, edges capture incumbent scientists who publish in the same venue repeatedly and we group interactions between scientists into "First-time" and "Repeated" categories.
%: \ie two incumbent scientists may have both appeared in the training data but collaborate for the \textit{first time} in the testing period or \textit{repeat} a previous collaboration.
%
In the inductive setting, there %is at least one "unseen" node in the test edges.
%There are then 
are 
three groups of interactions: "Unseen-to-Seen", "Unseen-to-Unseen", and "Seen-to-Unseen". For example, a graduate student ("unseen") can publish her first paper in the ACL community with her mentor ("seen"), or a group of scientists may publish in the ACL community for the first time ("Unseen-to-Unseen"). We distinguish "Unseen-to-Seen" and "Seen-to-Unseen" interactions between a "seen" and an "unseen" node input to the model as the directionality makes these two variants different in nature and predictive context; \eg a new graduate student may interact only with their mentor.
% while the mentor may publish with other new researchers.
%Finally, there can be new interactions between two "unseen" nodes in the testing period. For example, a group of scientists may publish in the ACL community for the first time.
%
We use these edge groups to understand the complexity of temporal link prediction task in both %transductive and inductive 
settings.
%We chronologically split the graphs into training and testing graph subsets.
%We assume the graphs are chronologically split into training and testing graph subsets.
% In this setup, there are nodes seen during the training data may present in the test data.
% This is natural as scientists (incumbents) may repeatedly publish in the same venue and other scientists (rookies) could publish for the first time.
We define the task complexity in the transductive setting as the proportion of "First-time" interactions with respect to the "Repeated" interactions.
% The higher the proportion, the higher the complexity of the transductive prediction task.
We take the proportion of "Seen-to-Unseen", "Unseen-to-Seen" and "Unseen-to-Unseen" interactions with respect to other edges to define the complexity of the inductive prediction task. 
The higher the proportion, the higher the complexity of both transductive and inductive tasks.
As shown in Figure~\ref{fig:task_complexity}, we notice that the complexity of the transductive task increases over different train/test splits over time (for both ACL and WoS graphs, there are emerging interactions between incumbent scientists) while inductive task complexity remains comparable.
While there are many new authors who publish over time in the ACL data, they are relatively low in the WoS data.

\subsection{Design of Evaluation}
% Previous works consider link prediction task as variants of the ranking and classification tasks.
% In a ranking solution, the model ranks the potential nodes that would interact with a given node in a future event.
% It assumes that a part of ground truth is available in the inference.
In this section, %we discuss evaluating the performance of the predictive models in the temporal link prediction.
% As we discussed previously, we narrow down the solution space into the variants of the ranking and classification models.
% \subsection{Proposing New Benchmark Tasks}
% \paragraph{\textbf{New Evaluation Procedures}}
we propose a systematic approach to improve the current evaluation procedures.
%
% \subsection{Temporal Link Prediction}
% \label{sec:task_description}
Previous work considers multi-step graph forecasting as a variant of the temporal link prediction task where interactions are either ranked or classified~\cite{visin2015renet,rossi2020temporal}.
% The aim is to predict the existence of a quadruple (\textit{$N_s$}, r, \textit{$N_o$}, t).
%They treat the temporal link prediction task as variants of the ranking and classification tasks.
In a ranking solution, the model ranks the potential nodes ($N_o$) that would interact with a given
node ($N_s$), relationship (r), and timestamp (t).
%
%In a ranking solution, the model ranks the potential nodes that would interact with a given node in a future event.
Metrics typically include mean rank (MR), mean reciprocal rank (MRR), and the percentage of examples with the  true target entity in top K candidates (known as Hits@K).
There are a few limitations in this approach.
First, we can only rank target nodes that are known (\ie seen in training data).
% This limits the ranking solutions to the transductive setting ("Seen-to-Seen") and a part of inductive setting ("Unseen-to-Seen").
Second, given a training graph with N nodes, we need to make $N^2$ inferences at most; 
This is intractable for very large graphs with millions of nodes.
Most recent works attempt to rank only a subset of target nodes ($M<<N$) to reduce the number of inferences ($M\times N <<< N^2$).
% This approach assumes the nodes seen in the training data may appear in the testing data as well.
However, performance heavily depends on the chosen subset of target nodes.
When treated as a classification task, models predict the existence of an edge
(\textit{$N_s$}, r, \textit{$N_o$}) at future time t.
%In a classification solution, the model predicts a probability for a pair of nodes in the test data.
Metrics include precision, recall, and AUC (area under the receiver operating characteristic curve).
This approach would be suitable for both transductive and inductive prediction tasks as its objective is to distinguish interactions from the non-interactions.
A common approach is to construct a set of negative examples equal to the number of positive examples in test sets, and thus performance can be heavily influenced by those chosen negative examples.

\paragraph{\textbf{Evaluation Recommendations}}
First, we propose to interpret the transductive and inductive task performance in context of our task complexity measure.
This indicates whether a model is capable of capturing the distribution shifts observed on the training and testing splits.
% For example, we noticed there are many emerging collaborations between rookie scientists in the ACL community.
% Since the graphs evolve over time, one might see 
Distribution shifts may create very different patterns of collaboration, partnership, and capability development in the testing period than in the training period.
% For example, there may be scientists who interact with emerging techniques.
However, detecting such a distribution shift or determining what causes a shift is hard~\cite{huyen2022designing}.
% It is even challenging to determine what causes a shift, or how the model should reacts to the changes in the patterns.
%The major challenge is to build models that are more robust to distribution (temporal) shifts.

Second, we recommend to report link prediction performance across multiple edge types.
For example, a model may perform well predicting \textit{collaboration} edges, but may not perform comparably predicting the \textit{scientist to capability} edges. This nuanced evaluation feedback can be used to target model improvement to boost overall performance or generalizability.
%
%Third, link 
Link prediction tasks should also be evaluated separately for the transductive and inductive settings, as discussed in Section~\ref{sec:problem_complexity}.
% We proposed two edge groups in the transductive setting and three edge categories in the inductive setting.
We notice that many recent works filter the test edges that co-appear in the train, valid, or test sets in the evaluation.
While the intention may be to focus on performance for unseen relationships, this is operationally irrelevant.
For example, one may be interested in predicting whether a group of scientists would repeatedly publish on the same conference.
These groups are persistent as they would appear on both train and test splits.
% On the other hand, they may build expertise on different subject areas over time.

% Fourth, we noticed that both ranking and classification solutions have caveats in the evaluation mainly due to the selection of target nodes (ranking) and negative samples (classification).
% Previous works perform the evaluation multiple times with different samples of target nodes (or negative samples) and report the average.
% % While this is an on-going research task, we propose few alternative directions to perform additional evaluation steps.
% Another approach is to develop classification methods to distinguish "First-time" and "Repeated" interactions.
% This method does not need to sample negative samples but distinguish the two subsets of the test edges.

%Fourth, 
Third,
nodes that do not receive updates regularly over time need to be explicitly accounted for. % in evaluation.
% For example, we record scientist-to-scientist interactions in our traces through co-authored publications.
There may be scientists who do not publish regularly in the same venue.
In this case, nodes may include gaps of activity in training data.
Predicting for these inactive/inconsistent scientists would be more challenging due to thee unusual activity flow compared to active scientists.
We need to design systems that would address this staleness problem.

Finally, our new datasets can support link prediction across multiple future time steps to enable evaluation of performance differences over time and with varying prediction windows.
%we can predict the links over multiple, future time steps rather than only in the next time step.
% This may allow the model to generalize over unseen entities in the training data who may interact with the events in the distant future.
%We can test how well our models predict future events, and how its ability changes with the number of future time steps.
There may be a drop in the performance with increasing time steps, or increasing gaps between train and test periods. % if models do not use any part of ground truth during the inference.

%% file: images/complexity_plot.tex
% Read data from csv into \data:
\pgfplotstableread[col sep=comma]{images/complexity_scores_merged.csv}\df
\small

%\vspace{-0.5\baselineskip}
\begin{tikzpicture} 
\begin{axis}[ 
    ybar,
   title= \textit{ACL} , % specify title, comment out for no title
   title style = {yshift=-1.75\baselineskip, font=\small}, 
   %xlabel=Year,% Specify Label for X-Axis, comment out for no label
    ylabel=Task\\Complexity,% Specify Label for X-Axis, comment out for no label
    ylabel style={yshift=-0.25\baselineskip, align=center, font=\small},
    height=1.1in, % height of plot
    width=2.9in,  %.25in,  % width of plot
    %xmin=0,xmax=24,% x-axis limits
    ymin=0,ymax=1,% y-axis limits
    xtick pos=left,
    tick align=center,
    ytick={0,0.25,0.5,0.75,1},
    xtick= {0,1,2,3,4},% xticks (can specify the yticks similarly)
    xticklabels={2016,2017,2018,2019,2020},%\empty,% hide xtick labels on top plots
    scaled x ticks=false, % don't scale ticks
    scaled y ticks=false, % don't scale ticks
    %every y tick label/.append style={font=\tiny} % makes all yticks be "tiny" fontsize
%legend style={fill=none, at={(0.5,-0.5)},%0.5)},
%	anchor=north,legend columns=2,column sep=.2cm,draw=none}, 
    ] 
     
    % ACL
    % transductive 
	 \addplot+[blue, no marks, solid,  thick] 
	 table [col sep=comma, x=index, y=ACLTransductive]\df; 
%\addlegendentry{ACL Transductive};
	 
    % inductive
	 \addplot+[orange, no marks, solid, thick] 
	 table [col sep=comma,x=index, y=ACLInductive]\df; 
%\addlegendentry{ACL Inductive};
	  
\end{axis}

 \end{tikzpicture}
 
\vspace{-0.5\baselineskip}

\begin{tikzpicture} 
\begin{axis}[ 
    ybar,
   title= WoS, % specify title, comment out for no title
   title style = {yshift=-1.75\baselineskip, font=\small}, 
   %xlabel=Year,% Specify Label for X-Axis, comment out for no label
    ylabel={Task\\Complexity},% Specify Label for X-Axis, comment out for no label
    ylabel style={yshift=-0.25\baselineskip, align=center, font=\small},
    height=1.1in, % height of plot
    width=2.9in,  %.25in,  % width of plot
    %xmin=0,xmax=24,% x-axis limits
    ymin=0,ymax=1,% y-axis limits
    xtick pos=left,
    tick align=center,
    ytick={0,0.25,0.5,0.75,1},
    xtick= {0,1,2,3,4},% xticks (can specify the yticks similarly)
    xticklabels={2016,2017,2018,2019,2020},%\empty,% hide xtick labels on top plots
    scaled x ticks=false, % don't scale ticks
    scaled y ticks=false, % don't scale ticks
    %every y tick label/.append style={font=\tiny} % makes all yticks be "tiny" fontsize
legend style={fill=none, at={(0.5,-0.3)},%0.5)},
	anchor=north,legend columns=2,column sep=.2cm,draw=none}, 
    ] 
    
    % wos
    % transductive 
	 \addplot+[blue, no marks, solid,  thick] 
	 table [col sep=comma, x=index, y=WOSTransductive]\df; 
\addlegendentry{Transductive};
	 
    % inductive
	 \addplot+[orange, no marks, solid, thick] 
	 table [col sep=comma,x=index, y=WOSInductive]\df; 
\addlegendentry{Inductive};

\end{axis}
 \end{tikzpicture}
  %\vspace{-1\baselineskip}

 \ignore{ 
\begin{tikzpicture} 
\begin{axis}[ 
    ybar,
   % title= , % specify title, comment out for no title
   % title style = {yshift=-0.75\baselineskip, font=\small}, 
   xlabel=Year,% Specify Label for X-Axis, comment out for no label
    ylabel=Task Complexity,% Specify Label for X-Axis, comment out for no label
    ylabel style={yshift=-0.25\baselineskip, align=center, font=\small},
    height=1.25in, % height of plot
    width=2.9in,  %.25in,  % width of plot
    %xmin=0,xmax=24,% x-axis limits
    %ymin=0,ymax=20,% y-axis limits
    xtick pos=left,
    tick align=center,
    xtick= {0,1,2,3,4},% xticks (can specify the yticks similarly)
    xticklabels={2016,2017,2018,2019,2020},%\empty,% hide xtick labels on top plots
    scaled x ticks=false, % don't scale ticks
    scaled y ticks=false, % don't scale ticks
    %every y tick label/.append style={font=\tiny} % makes all yticks be "tiny" fontsize
legend style={fill=none, at={(0.5,-0.5)},%0.5)},
	anchor=north,legend columns=2,column sep=.2cm,draw=none}, 
    ] 
    
    % wos
    % transductive 
	 \addplot+[blue, no marks, solid,  thick] 
	 table [col sep=comma, x=index, y=WOSTransductive]\df; 
\addlegendentry{WoS Transductive};
	 
    % inductive
	 \addplot+[blue, no marks, dashed, thick] 
	 table [col sep=comma,x=index, y=WOSInductive]\df; 
\addlegendentry{WoS Inductive};
	 
    % ACL
    % transductive 
	 \addplot+[orange, no marks, solid,  thick] 
	 table [col sep=comma, x=index, y=ACLTransductive]\df; 
\addlegendentry{ACL Transductive};
	 
    % inductive
	 \addplot+[orange, no marks, dashed, thick] 
	 table [col sep=comma,x=index, y=ACLInductive]\df; 
\addlegendentry{ACL Inductive};
	  
\end{axis}
 \end{tikzpicture}
  }

%% file: sections/discussion.tex
\section{Summary and Conclusions}% and Discussion}
In this paper, we release seven \dhgraphs benchmark datasets 
% that support numerous tasks of interest 
to understand how scientific collaboration, partnership and authorship evolve in AI and nuclear nonproliferation communities.
These graph datasets consist of ~3.5M nodes and ~34M timestamped edges in total and we
% We establish temporal link prediction tasks to forecast collaboration, partnership and authorship patterns.
%We 
% group the edges into multiple categories based on the co-appearance in the training and testing data and 
show the complexity of transductive and inductive tasks through a systematic approach.
We hope our contributions will help researchers to build and evaluate new graph models, or understand the limitations of existing graph models in  dynamic heterogeneous graph forecasting tasks.